\documentclass[conference]{ieeetran}
\usepackage{amsmath,graphicx,color,algorithm,algorithmic,subfigure}
\usepackage{flushend,array}
\usepackage[usenames,dvipsnames,table]{xcolor}

\usepackage{fancyhdr}
\title{Medical Image Classification via SVM\\ using LBP Features from Saliency-Based Folded Data}

\author{\authorblockN{Zehra \c{C}amlica\authorrefmark{1}, H.R. Tizhoosh\authorrefmark{1} and Farzad Khalvati\authorrefmark{2} }
\authorblockA{\authorrefmark{1} KIMIA Lab, University of Waterloo, Canada, [\emph{tizhoosh@uwaterloo.ca}]}
\authorblockA{\authorrefmark{2} Sunnybrook Health Sciences Centre, Toronto, Canada
}
}

\begin{document}
\maketitle
\thispagestyle{fancy}
\begin{abstract}
Good results on image classification and retrieval using support vector machines (SVM) with local binary patterns (LBPs) as features have been extensively reported in the literature where an entire image is retrieved or classified. In contrast, in medical imaging, not all parts of the image may be equally significant or relevant to the image retrieval application at hand. For instance, in lung x-ray image, the lung region may contain a tumour, hence being highly significant whereas the surrounding area does not contain significant information from medical diagnosis perspective. In this paper, we propose to detect salient regions of images during training and fold the data to reduce the effect of irrelevant regions. As a result, smaller image areas will be used for LBP features calculation and consequently classification by SVM. We use IRMA 2009 dataset with 14,410 x-ray images to verify the performance of the proposed approach. The results demonstrate the benefits of saliency-based folding approach that delivers comparable classification accuracies with state-of-the-art but exhibits lower computational cost and storage requirements, factors highly important for big data analytics.
\end{abstract}


\section{Introduction}
\label{sec:intro}
Recent advances in medical imaging devices have led to the generation of big image data on a daily basis. The main purpose of medical information system is the acquisition of necessary information to provide high-quality care through accurate and efficient diagnosis and treatment planning~\cite{2}.

In order to implement advanced information systems operating on large databases (hence handling big image data), suitable methods are required to respond to a query (an image selected by a clinician) by retrieving images that have similar characteristics. Content-based image retrieval (CBIR) uses image search techniques that incorporate visual features, such as color, texture, and shape, in order to respond to user's queries. In medical imaging context, CBIR can enormously contribute to more reliable diagnosis, among others, by classifying the query image and retrieving similar images already annotated by diagnostic descriptions and treatment results. 

The main purpose of this work is to obtain high classification score with less computational complexity and lower storage requirements. In order to save time and to gain high classification score, first a salient region detector is used~\cite{8}. Next, images are folded to mainly contain salient areas, and reduce the effect of irrelevant (non-salient) regions. Subsequently, we can extract LBP features from folded images and classify them via SVM. We use IRMA x-ray dataset with 14,410 images for training and testing. The classification result is computed with reported ImageCLEF error score evaluations for different methods \cite{18}.

\vspace{0.2in}
This paper is organized as follows: In section \ref{sec:BGreview}, a brief background review on medical image retrieval is given. In section \ref{sec:alg} we describe the proposed approach. Section \ref{sec:experiments} reports the experimental results using IRMA dataset. Section \ref{sec:concl} concludes the paper.

\section{Literature Review}
\label{sec:BGreview}
There is a clear demand for fast and accurate image search technologies in clinical settings when physicians (e.g. radiologists) desire to search for similar images of all patients in the past when examining a current patient. Content-based image retrieval (CBIR) has been subject to research to satisfy some aspects of this demand. CBIR takes advantage of visual contents of an image such as colour, shape and texture to search for (similar) images in large archives. Generally, a software system that can access medical archives to search for similar images is a CBIR system. The ``content-based'' aspect of CBIR simply means that the search is conducted based on some visual (pictorial) features of the image, and not based on text annotations (the latter is mainly used when we search on the internet). Some examples for medical CBIR systems are TELEMED \cite{3}, ASSERT \cite{4} and IRMA~\cite{5}. 

The features used in CBIR can be textual or visual. Recent medical image retrieval systems increasingly rely on visual features that could be low-level features (primitive), middle-level features (logical), and high-level features (abstract). Almost all early CBIR systems are based on low-level features (colour or shape), but recently, mid- and high-level image representations have received more attention. Mid-level features are obtained from particular parts of the image, which are important regions with significant details~\cite{6,7,8,9}. High-level features are represented with semantic design. The semantic design (high-level features such as emotions, objects and events) can be present in visual or textual information. 

Local binary patterns (LBP) are utilized as features for texture description \cite{12}. LBP descriptors are commonly used in facial expression analysis and recognition \cite{13,14,15}. LBP measures invariant texture of gray-scale images with utilization of local neighborhoods. The basic LBP operator replaces pixel values with labels by binarizing $3\times 3$ neighborhoods around each pixel with the centre pixel as a threshold. Pixel labels are then converted to decimal numbers. Because LBP is an easy-to-compute feature extraction method, it has been successfully used in many studies such as face recognition and image annotation~\cite{13,19,20,21,22}. In the proposed method, LBP is applied to multi-block patches in the image at different scales. After labeling the image parts, the feature histogram is extracted from the local region labels. The regions can be rectangular, circular or triangular. Recently, a new approach to binary encoding of local image information is proposed which uses ``barcodes'' based on thresholding projections via Radon transform \cite{Tizhoosh2015}. \\
\vspace{0.2in}
Different methods can be used to classify images \cite{Wang2010, Othman2013,Bosch2007,Arvacheh2005}. For our classification, we use SVM in this paper, which is a supervised learning method to classify datasets. It investigates sets of feature vectors in an $N$ dimensional space. It uses support vectors to construct a hyperplane to separate different classes by maximizing the margin between them defined by the given hyperplane~\cite{25}. 

\section{The Proposed Approach}
\label{sec:alg}
In this section, we present the proposed image classification method. This approach comprises of a pre-processing phase,  offline training and an online usage phase. During pre-processing, saliency maps are extracted and images are folded. SVM is trained using LBP features of both folded and not folded images in the offline phase. Finally, online classification is described. Algorithm \ref{alg:approach} gives a generic overview of the proposed approach.

\begin{algorithm}[tb]
\caption{Proposed Approach}
\begin{algorithmic}[1]
\label{alg:approach}
\STATE ------- Pre-Processing -------
\STATE Read all images $I_i$
\STATE Calculate saliency template $S^*$ 
\STATE $I_i^F \leftarrow$ Apply folding on all images $I_i$
\STATE Save $S^*$, and all folded images $I_i^F$ 
\STATE ------- Training -------
\STATE Read folded images $I_i^F$ 
\STATE Set number of classes $N_C$
\STATE Extract LBP features from folded data
\STATE Train SVM to generate the support vectors $\mathbf{v}_1,\mathbf{v}_2,\dots$ 
\STATE Save $\mathbf{v}_1,\mathbf{v}_2,\dots$
\STATE ------- Online Classification -------
\STATE Read the query image $I^q$
\STATE Read the saliency template $S^*$
\STATE Read the support vectors $\mathbf{v}_1,\mathbf{v}_2,\dots$
\STATE $I^q_S \leftarrow$ Apply the saliency template $S^*$ on $I^q$ 
\STATE $I^q_F \leftarrow$ Apply folding $F$ on $I^q_S$
\STATE Extract LBP features from $I^q_F$
\STATE Classify the query using SVM
\end{algorithmic}
\end{algorithm}

\subsection{Preprocessing}
The pre-processing of image data mainly consists of two procedures. The first procedure creates a saliency template, and the second procedure formulates the image folding based on the saliency template.

\subsubsection{Saliency Map} 
The detection of salient regions of an image is crucial to extract effective information. We propose to create a saliency template by averaging all saliency maps which are detected by context-aware saliency algorithm~\cite{8}. 

The context-aware saliency algorithm detects image regions that best represent the ``scene''. It is a detection algorithm, as its authors state, ``based on four principles observed in the psychological literature: local low-level considerations, global considerations, visual organizational rules, and
high-level factors''. Local low-level factors (such as contrast and color), global calculations suppressing frequently-occurring features, visual organization rules (visual forms may possess one or several centres of gravity) and high-level factors (such as priors on the salient object location and object detection) are considered by the algorithm. The implementation of this algorithm is available on the authors' website\footnote{http://webee.technion.ac.il/labs/cgm/Computer-Graphics-Multimedia/Software/Saliency/Saliency.html}.

\begin{figure}[h!]
\center
\includegraphics[width=0.9\columnwidth]{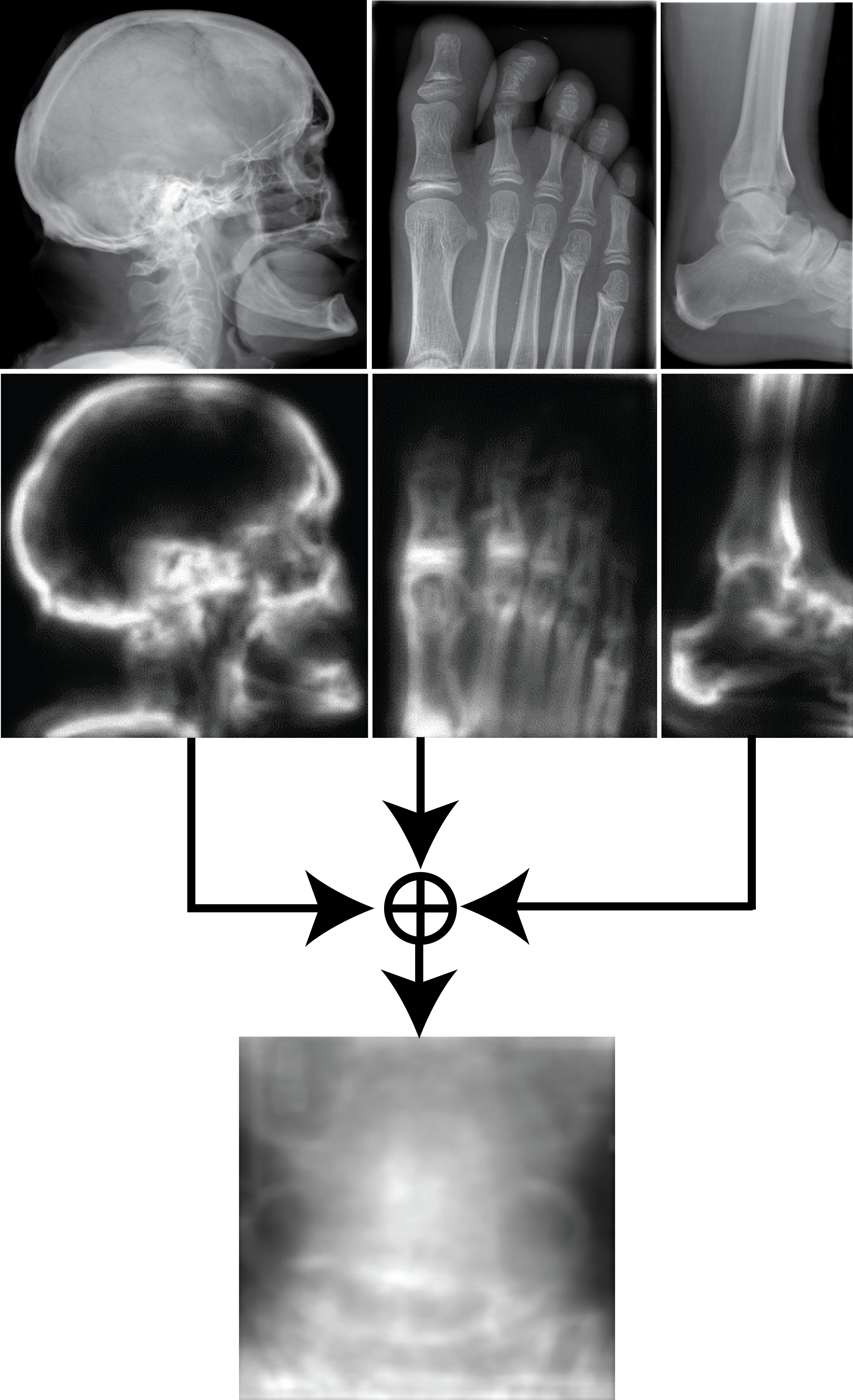}
\caption{A saliency map is generated for each available training image. A saliency template is then assembled by combining all saliency maps.}
\label{fig:res}
\end{figure}

Saliency maps of all training images are generated and averaged to calculate a saliency  template $S^*$ (Algorithm \ref{alg:saliency}). Figure \ref{fig:res} shows  three images, their saliency maps, and the saliency template created by averaging all saliency maps. The average of saliency maps is first  calculated internally within each class, then the average is taken across all classes.

\begin{algorithm}[tb]
\caption{Pre-Processing Stage: Saliency Template $S^*$}
\begin{algorithmic}[1]
\label{alg:saliency}
\STATE $N_C \leftarrow$ number of classes; $i=1$.
\STATE Initialize saliency template $S^*= []$
\WHILE{$i<N_C$}
	\STATE Calculate the saliency map $S_i$ for image $I_i$ \cite{8}
	\STATE $S^* \leftarrow S^* + S_i$
	\STATE $i \leftarrow i +1$
\ENDWHILE
\STATE $S^* \leftarrow \frac{S^*}{N_C}$
\end{algorithmic}
\end{algorithm}

The salient, less salient and not salient areas are defined for training data by dividing images to $N$ sub-blocks. Then, based on the saliency template, the folding is applied. The new images with reduced area can now be used for local pattern analysis. 

\subsubsection{Image Folding} 
\begin{figure*}[!htb]
\centerline{\includegraphics[width=0.7\textwidth]{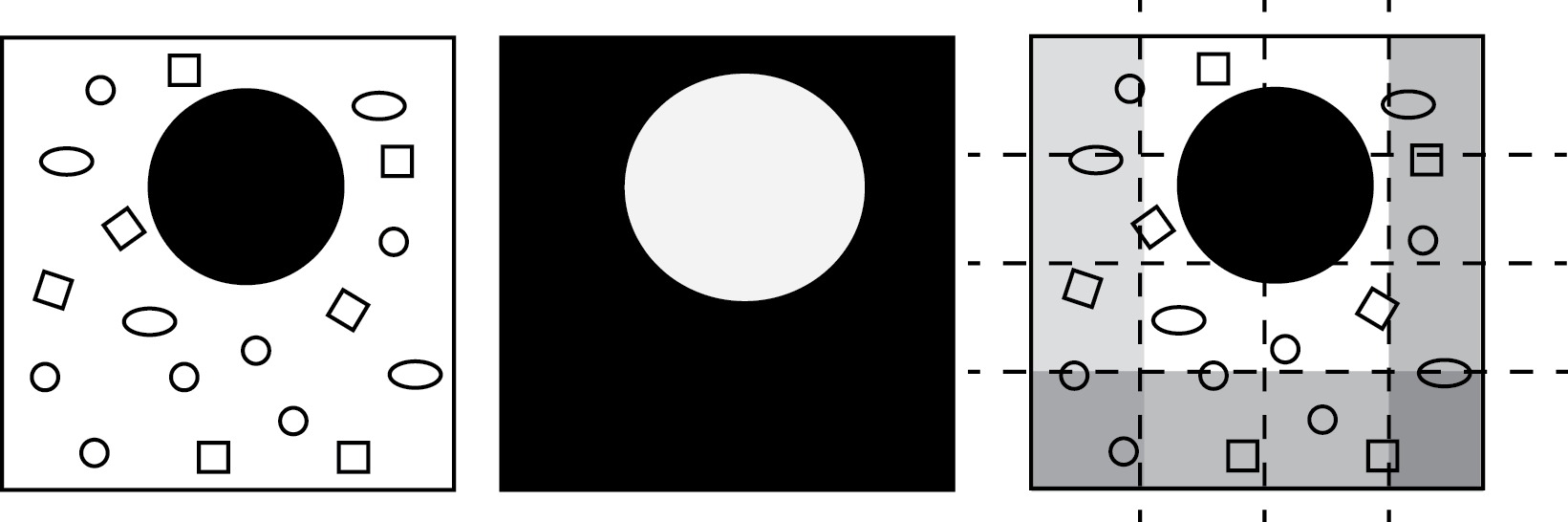}}
\caption{Schematic illustration of saliency maps and image folding: The input image (left image) is processed to find a salient region (middle image). Subsequently, non-salient regions (right image, gray stripes) are marked to be folded inwardly.}
\label{fig:Saliency_Folding}
\end{figure*}

Folding the rectangular region $A\subset I$ within image $I$ resulting in an image $I'\subset I$ can be given through $I' = A + I \backslash A$ whereas the sign ``$\backslash$'' denotes the set-theoretical subtraction. The main purpose of folding is to reduce image area without loosing information but reducing the dimensionality of features (see Fig. \ref{fig:Saliency_Folding}). The folding steps are described in Algorithm \ref{alg:folding}.

\begin{algorithm}[htbp]
\caption{Pre-Processing Stage: Image Folding}
\begin{algorithmic}[1]
\label{alg:folding}
\STATE Set number of blocks $M$ ($=N\times N= 4\times 4$) 
\STATE Read saliency tempalte $S^*$
\STATE Read the input image $I$
\WHILE{not all combinations tested} 
	\STATE Align two columns 
	\STATE Take the summation of all pixel values in $S^*$
	\STATE Keep $s_{\textrm{max}}^{\textrm{c}_i}$  (maximum value of summed columns) 
	\STATE Update $s_{\textrm{max}}^{\textrm{column}} \leftarrow \sum_i s_{\textrm{max}}^{\textrm{c}_i}$.
\ENDWHILE
\WHILE{not all combinations tested} 
	\STATE Align two rows 
	\STATE Take the summation of all pixel values in $S^*$
	\STATE Keep $s_{\textrm{max}}^{\textrm{r}_j}$  (maximum value of summed rows) 
	\STATE Update $s_{\textrm{max}}^{\textrm{row}} \leftarrow \sum_i s_{\textrm{max}}^{\textrm{r}_j}$.
\ENDWHILE
\STATE Find the folding $F_{\textrm{best}}$ that satisfies
\STATE  \qquad \qquad$s = \min(s_{\textrm{max}}^{\textrm{column}},s_{\textrm{max}}^{\textrm{row}})$.
\STATE Apply the folding $F_{\textrm{best}}$ to $I$.
\end{algorithmic}
\end{algorithm}

\subsection{Offline Training}
LBP features are extracted from $K$ ($M>K$) divided sub-blocks of image with different scaling factors (1 and 2). LBP feature vector for an image has 1,062 dimensions with the following condition: $M=4\times 4, K=3\times 3$. The LBP histogram features from  training data are used to train multi-class SVM \cite{27} to classify images. The SVM kernel type is set to be Radial Basis Function. 

\subsection{Online Classification}
In online part, an image query is selected from IRMA~\cite{5} test database and LBP features are calculated for the saliency-based folded image as new images are encountered. Next, SVM classification is performed with LBP features. We also run the experiments for the LBP-SVM without folding.

\section{Experiments and Results}
\label{sec:experiments}

\subsection{Data Set}
The Image Retrieval in Medical Applications (IRMA\footnote{http://irma-project.org/}) database is a collection of 14,410 x-ray images that have been randomly collected from daily routine work at the Department of Diagnostic Radiology of the RWTH Aachen University\footnote{http://www.rad.rwth-aachen.de/}.  The downscaled images were collected from different ages, genders, view positions, and pathologies~\cite{18}. 

Each image in the dataset has an IRMA code. According to these codes, 193 classes are defined. The IRMA code comprises four axes with three to four positions each: 1) the technical code (T) (modality), 2) the directional code (D) (body orientations), 3) the anatomical code (A) (body region), and 4) the biological code (B) (the biological system examined). The complete IRMA code consists of 13 characters TTTT-DDD-AAA-BBB, with each character in $\{0,\dots,9; a,\dots,z\}$. As many as 12,677 images are separated for training. The remaining 1,733 images are used as test data.

\begin{figure*}[htb]
\centering     
\subfigure[1121-127-700-500]{\label{fig:a}\includegraphics[width=30mm,height=30mm]{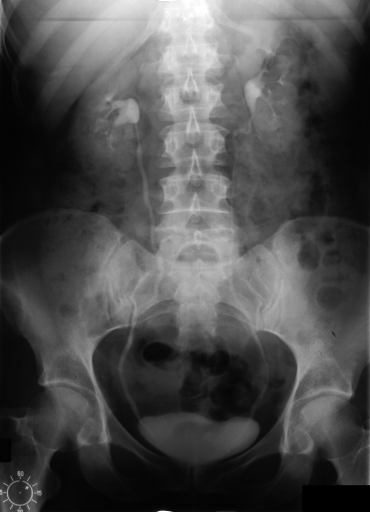}}
\subfigure[1121-120-918-700]{\label{fig:b}\includegraphics[width=30mm,height=30mm]{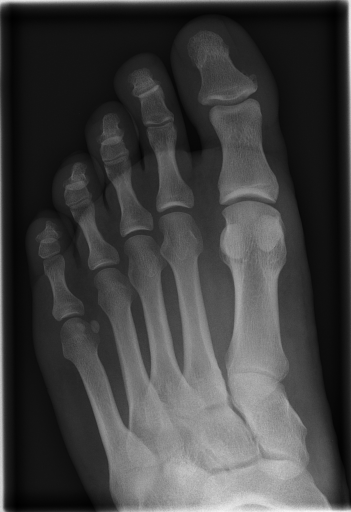}}
\subfigure[1121-120-942-700]{\label{fig:b}\includegraphics[width=30mm,height=30mm]{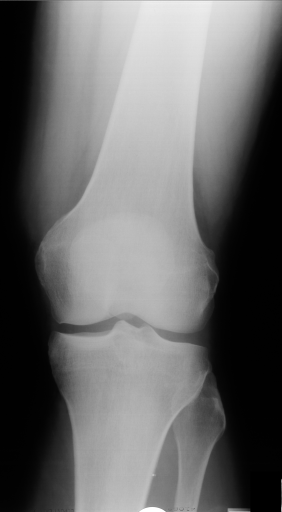}}
\subfigure[112d-121-500-000]{\label{fig:b}\includegraphics[width=30mm,height=30mm]{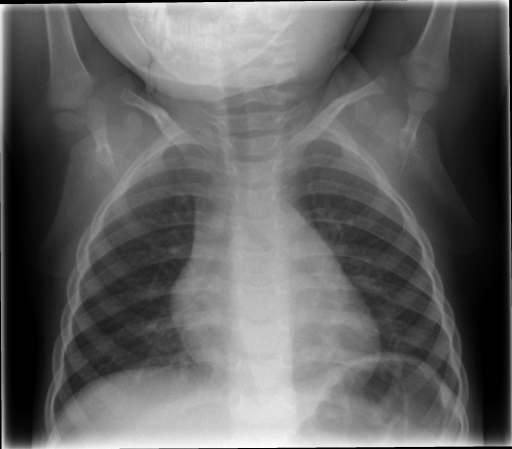}}
\subfigure[1123-127-500-000]{\label{fig:b}\includegraphics[width=30mm,height=30mm]{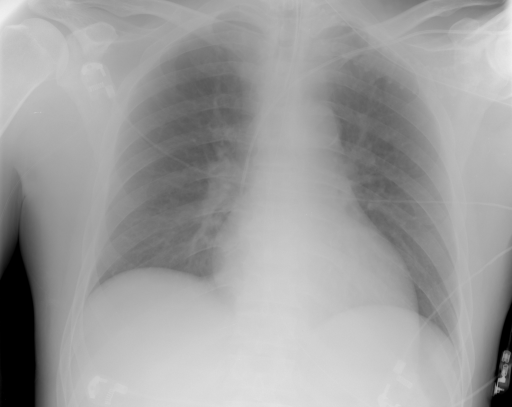}}\\
\subfigure[1121-120-200-700]{\label{fig:b}\includegraphics[width=30mm,height=30mm]{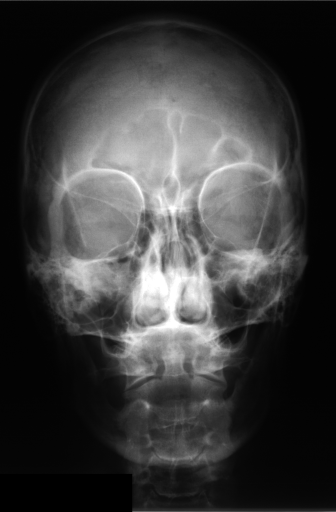}}
\subfigure[1121-200-412-700]{\label{fig:b}\includegraphics[width=30mm,height=30mm]{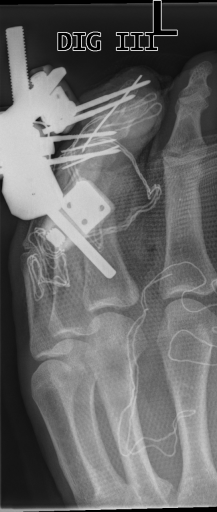}}
\subfigure[1121-110-414-700]{\label{fig:b}\includegraphics[width=30mm,height=30mm]{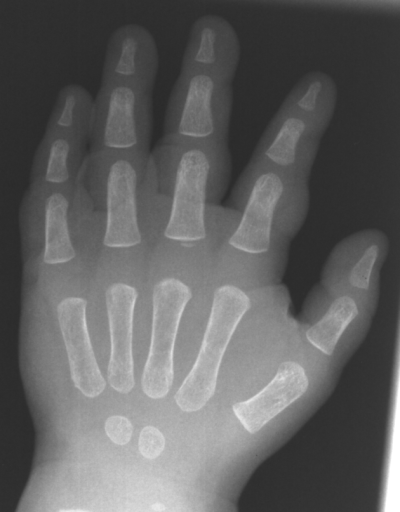}}
\subfigure[1121-240-442-700]{\label{fig:b}\includegraphics[width=30mm,height=30mm]{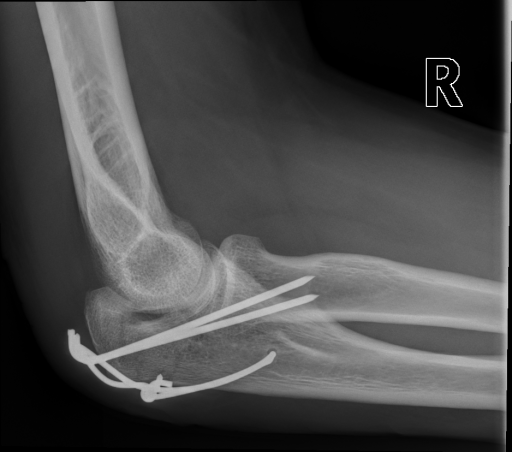}}
\subfigure[1121-220-310-700]{\label{fig:b}\includegraphics[width=30mm,height=30mm]{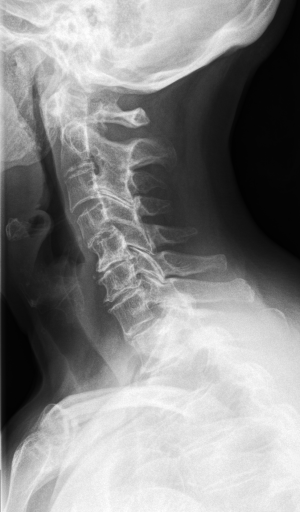}}

\caption{Sample images from IRMA Dataset with their IRMA codes TTTT-DDD-AAA-BBB.}
\label{fig:IRMASamples}
\end{figure*}

Figure \ref{fig:IRMASamples} shows some samples images from IRMA dates along with their corresponding IRMA codes. 

\subsection{Error Measurement} 
The ImageCLEF project has defined an error score evaluation method in order to evaluate the classification performance of methods on IRMA dataset \cite{18}. As in IRMA dataset all images are labelled with the technical, directional, anatomical and biological independent axes, the error $E$ can be defined as follows

\begin{equation}
E = \sum\limits_{i=1}^n\dfrac{1}{b_{i}}\dfrac{1}{i}\delta(I_{i},\hat{I_{i}})
\label{6}
\end{equation}

 where $b_{i}$ is number of possible labels at position $i$ and $\delta$ is the decision function delivering $1$ for wrong label and $0$ for correct label when the IRMA codes of the image $I_i$ is compared with the IRMA code of the image $\hat{I}_i$. For every axis, the maximal possible error is computed and the errors are normalized between $0.25$ and $0$. If all positions in all axes are wrong, error value is $1$.

\subsection{Classification Error} 
The experiments resulted in an error score of $153.07$ for the proposed method of SVM image classification with multi-scale LBP on saliency-based folded image. If images are not folded, the SVM  error slightly decreases to $146.55$. This slight decrease in error comes with a higher cost in computation; the dimensions of features are twice the dimensions of the folded image. This means that the accuracy does not fall while time and computational cost are decreasing. Saliency-based folding reduces complexity without loosing important patterns in salient region. The computational complexity decreases because folding reduces the feature vector dimension. 
 
 Without consideration of salient area, folding was tried in different directions. The error is clearly  increased. apparently, saliency template plays a crucial role in deciding how to fold an image. 
 
 For sake of comparison, the IRMA dataset was used in ImageCLEF 2009 competition with 2008 IRMA code  and basic LBP with $4\times 4$ multi-blocks is applied in \cite{22} and the error score is reported as $261.2$ \cite{22}. In addition, the lowest error score in ImageCLEF 2009  with 2008 IRMA code is $169.5$ \cite{18}. The comparison of classifiers and SVM results are outlined in Table~\ref{table1}. 

\subsection{Memory and Time} 
The image area is reduced by $50\%$ with saliency-based folding. As an effect, the number of feature dimension decreased from 1,888 to 1,062 which equals $44\%$ decrease in feature dimensionality. 

SVM needs $141.17$ seconds training time and $92.51$ seconds testing time without saliency-based folding. That corresponds to 53 milliseconds per image for online queries. 

In contrast,  with saliency-based folding SVM only needs $60.36$ seconds training time and $52.56$ seconds testing time. That corresponds to $30$ milliseconds per image for online queries. To neglect the overhead for the saliency calculations, and only by looking at the testing times (online execution), using the proposed approach accelerates the classification process by roughly $43\%$ when looking at online computation times per query.

\vspace{0.1in}
\begin{table}[h]
\centering
\setlength\extrarowheight{5.5pt}
\begin{tabular}{lcc}   
Method & Error & $t$ (ms)/image\\ \hline \hline 
MS$^{4\times 4}$ LBP/SVM & $146.55$ & $53$\\ \hline 
MS$^{3\times 3}$ LBP/SVM w. folding & $153.07$	& $30$\\ \hline
TAU \cite{18}   & $169.5$ & - \\ \hline
VPASabanci \cite{22}   		& $261.2$ & -\\
 \hline \hline
\end{tabular}
\caption{Image classification results (MS$^{n\!\times\!n}\!=\! n\!\times\! n$ multi scale, $t$= time), results of TAU and VPASabanci as reported in literature.}
\label{table1}
\end{table}%

\section{Conclusions}
\label{sec:concl}
Content-based image retrieval (CBIR) depends on good classification first to assign a query to a the right image category. The time requirements become paramount hone dealing with big data.  

The proposed medical image classification using saliency-based folding method appears to be an effective method when support vector machines and local binary patterns are employed. Folding non-salient (non-relevant)  parts of the image may result in slight increase of classification error. That may be expected since folding areas overlap with salient regions resulting in slight distortion. However, the proposed approach does accelerate the online classification, an advantage that might be crucial for big image data (reduction from $53$ millisecond per image to $30$ milliseconds corresponding to $43\%$ acceleration). 

The decision how to fold image blocks is the most critical part of the pre-processing. Different approaches can be examined in future work to investigate the feasibility and the potential effect of folding blocks and not necessarily just folding rows and columns. As well, one may consider the deletion of non-salient blocks altogether. This may be particularly of interest in non-medical cases where the scene may contain irrelevant information along with objects of interest.

\vspace{0.2in}
As a potential future work, one may also investigate the incorporation of the new barcode technology \cite{Tizhoosh2015} into retrieval-oriented classification combined with optimization techniques that employ the concept of opposite entities \cite{Tizhoosh2007a, Tizhoosh2008,Ventresca2007}.

\bibliographystyle{IEEEbib}
\bibliography{reff}

\end{document}